\documentclass[10pt,journal,compsoc]{IEEEtran}

\usepackage{caption}
\usepackage[pdftex]{graphicx}
\graphicspath{{./Figures/}}
\ifCLASSOPTIONcompsoc
\usepackage[caption=false,font=footnotesize,labelfont=sf,textfont=sf,subrefformat=parens,labelformat=parens]{subfig}
\else
\usepackage[caption=false,font=footnotesize,subrefformat=parens,labelformat=parens]{subfig}
\fi
\usepackage{soul}

\usepackage{hyperref}
\usepackage{amsmath}
\usepackage{amssymb}
\usepackage{algorithm}
\usepackage{algpseudocode}

\usepackage{booktabs}
\usepackage{multicol}
\usepackage{multirow}
\usepackage[dvipsnames,table]{xcolor}
\usepackage{tikz}
\usepackage{etoolbox}
\definecolor{tabletopheader}{gray}{.6}
\colorlet{tableheader}{TealBlue}
\definecolor{tableoddrow}{gray}{.95}
\definecolor{tableevenrow}{gray}{.9}
\renewcommand{\arraystretch}{1.25} 
\newcolumntype{L}{>{\ifnumequal{\rownum}{1}{\sffamily\color{white}}}l}
\newcolumntype{C}{>{\ifnumequal{\rownum}{1}{\sffamily\color{white}}}c}
\newcolumntype{R}{>{\ifnumequal{\rownum}{1}{\sffamily\color{white}}}r}

\newcommand{\etal}{\textit{et~al.}}

\newcommand{\Title}{Identifying the Key Attributes in an Unlabeled Event Log for Automated Process Discovery}

\begin{document}
\title{\Title}
\author{Kentaroh~Toyoda,~\IEEEmembership{Member,~IEEE}, %
        Rachel Gan Kai Ying,
        Allan NengSheng Zhang, and
        Tan Puay Siew
        \thanks{
            Kentaroh Toyoda is with the Institute of High Performance Computing (IHPC), Agency for Science, Technology and Research (A*STAR), 1 Fusionopolis Way, \#16-16 Connexis, Singapore 138632, Republic of Singapore.
            E-mail: \url{kentaroh.toyoda@ieee.org}\protect
        }
        \thanks{Rachel Gan Kai Ying is with Nanyang Technological University (NTU), Republic of Singapore.}
        \thanks{Allan NengSheng Zhang, and Tan Puay Siew are with Singapore Institute of Manufacturing Technology (SIMTech), Agency for Science, Technology and Research (A*STAR), 5
CleanTech Loop \#01-01, CleanTech Two Block B, Singapore 636732, Republic of Singapore.}
        \thanks{The research was conducted when K. Toyoda and Gan K. Y. R. were in SIMTech.}
}

\markboth{IEEE Transactions on Services Computing,~Vol.~XX, No.~X, Xxxxxx~XXXX}%
{Toyoda \MakeLowercase{\etal}: \Title}

\maketitle
\begin{abstract}
Process mining discovers and analyzes a process model from historical event logs. The prior art methods use the key attributes of case-id, activity, and timestamp hidden in an event log as clues to discover a process model. However, a user needs to specify them manually, and this can be an exhaustive task. 
In this paper, we propose a two-stage key attribute identification method to avoid such a manual investigation, and thus this is a step toward fully automated process discovery. One of the challenging tasks is how to avoid exhaustive computation due to combinatorial explosion. For this, we narrow down candidates for each key attribute by using supervised machine learning in the first stage and identify the best combination of the key attributes by discovering process models and evaluating them in the second stage. Our computational complexity can be reduced from $\mathcal{O}(N^3)$ to $\mathcal{O}(k^3)$ where $N$ and $k$ are the numbers of columns and candidates we keep in the first stage, respectively, and usually $k$ is much smaller than $N$. We evaluated our method with 14 open datasets and showed that our method could identify the key attributes even with $k = 2$ for about 20 seconds for many datasets.
\end{abstract}

\IEEEpeerreviewmaketitle
\section{Introduction}
Process mining enables us to discover and analyze a business process model from historical event logs~\cite{Van_Der_Aalst2012-db}. In particular, process discovery, as its name suggests, is the step to discover a business process model by finding a pattern of sequential processes in an event log. Once a process model is obtained, we can use it for further analysis such as conformance check (i.e., identifying whether a newly given event log follows the discovered process model) and bottleneck analysis (i.e., identifying which parts of processes are bottlenecks and can be improved).
They have been proven to be successful in many industries, including supply chain (e.g.,~\cite{Jokonowo2018-nm}) and manufacturing (e.g.,~\cite{Mahendrawathi2018-po, Ruschel2020-pj, Lorenz2021-du}).

\begin{figure*}[t]
\begin{minipage}[t]{0.3\linewidth}
    \resizebox{\linewidth}{!}{
        \rowcolors{2}{tableoddrow}{tableevenrow}
        \centering
        \sffamily
        \begin{tabular}{CLLC}
            \rowcolor{tableheader}
            ID & Activity & Datetime & \dots \\
            1337 & Take Order & 01/04/2020 1:37PM & \dots \\
            1337 & Note Address & 01/04/2020 1:39PM & \dots \\
            1337 & Note Payment Method & 01/04/2020 1:40PM & \dots \\ 
            1337 & Grab Soda & 01/04/2020 1:42PM & \dots \\
            1337 & Prepare Burger & 01/04/2020 1:41PM & \dots \\
            1337 & Wrap Order & 01/04/2020 1:53PM & \dots \\
            1337 & Deliver Order & 01/04/2020 1:55PM & \dots \\
            1338 & Take Order & 01/04/2020 1:42PM & \dots \\
            \dots & \dots & \dots & \dots \\
        \end{tabular}
    }
\end{minipage}
$\rightarrow$
\begin{minipage}[c]{0.67\linewidth}
    \centering
    \includegraphics[width=\linewidth]{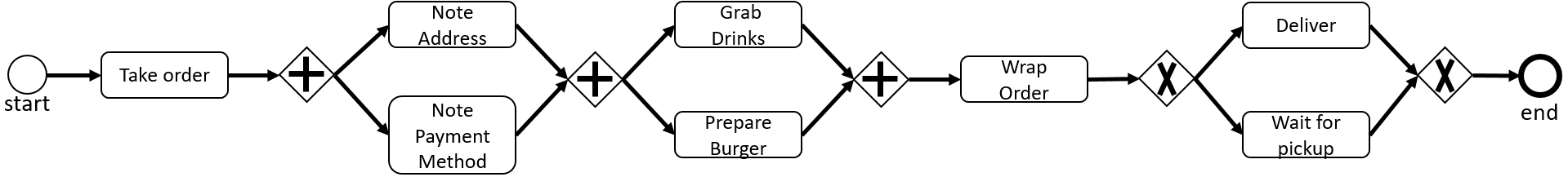}
\end{minipage}
\caption{\textbf{Left:} An example of event logs; \textbf{Right:} Process model discovered from the table.}%
\label{fig:process_model}
\end{figure*}

Process discovery algorithms, often called miners, have been well-studied in the last two decades (e.g.,~\cite{Cook1995-zg, Van_der_Aalst2002-xu, Weijters2006-yz, Leemans2013-iv, Vanden_Broucke2017-rp, Augusto2019-cb}). 
The basic idea of process discovery is to find the frequent patterns of ordered and parallel execution of processes.
Most of the existing algorithms discover a process model using three attributes in an event log, namely (i) case-id, (ii) timestamp, and (iii) activity (hereafter, the key attributes). Processes with the same case-id are a series of correlated processes. A timestamp is used in process discovery to know the order of processes in the same case-id, and activity is the name of a process. \figurename~\ref{fig:process_model} shows an example of an event log in a hamburger shop and its discovered process model.\footnote{We modified the example event log in \url{https://pm4py.fit.fraunhofer.de/getting-started-page}} In this example, the ID can be seen as the case-id as the processes in the same case-id seem to be correlated. Also, the activity and datetime columns are used as the activity and timestamp, respectively. 

Most of the existing process discovery algorithms require users to specify the key attributes in an event log. However, such a manual approach is not desirable when an event log contains dozens of or even hundreds of columns. Furthermore, as process mining may not necessarily be executed by practitioners, an automated process discovery where a user does not have to specify the key attributes in an event log would be appreciated in many situations.

In this line of work, Abbad Andaloussi~et~al. proposed a method to identify a case-id column in an event log~\cite{Abbad_Andaloussi2018-ur}. Their idea is to test every column in the event log as the case-id attribute and evaluate a process model discovered by such an assumption. However, this method only identifies the case-id attribute. Second, its computation is exhaustive as it repeatedly discovers and evaluates process models for every case-id candidate. If we were to identify case-id, timestamp, and activity attributes with their method, the computation cost would be $\mathcal{O}(N^3)$ where $N$ is the number of columns of a given event log.

This paper considers the scenario where raw data are given in a table format and the columns of case-ids, activity, and timestamp are hidden in the table. The objective is to identify these attributes efficiently. 
More specifically, let $L$ be a table-formated raw log data with $N_\text{rows}$ rows and $N_\text{attr}$ attributes, i.e., $A = \{a_1, a_2, \cdots, a_{N_\text{attr}}\}$ where $N_\text{attr} \geq 3$. Furthermore, $L$ consists of raw event values $E = \{\mathbf{e}_1, \mathbf{e}_2, \cdots, \mathbf{e}_{N_{\text{rows}}}\}$ where $\mathbf{e}_i = \{v_1^i, v_2^i, \cdots, v_{N_\text{attr}}^i\}$. The objective is to identify (i) case-id, (ii) activity, and (iii) timestamp attributes in $A$ by analyzing $E$. In this setting, we propose a two-stage key attribute identification method for automated process discovery. To avoid exhaustive computation due to combinatorial explosion, we first narrow down candidates for each key attribute by using supervised machine learning. If we could identify a single candidate for each key attribute, then we output them as the key attributes. Otherwise, we identify the best combination of the candidates by discovering and evaluating a process model for every combination. We define five scoring functions for this evaluation. Our computational complexity can be reduced from $\mathcal{O}(N^3)$ to $\mathcal{O}(k^3)$ where $k$ is the number of candidates we keep in the first stage, where usually $k$ is much smaller than $N$. We thoroughly evaluated our method with 14 open datasets, clarified which parameters affect accuracy and computation, and showed that our method could feasibly identify the key attributes even with $k = 2$ within a reasonable time (about 20 seconds).

The contributions of our research are as follows:
\begin{enumerate}
    \item To the best of our knowledge, we are the first to identify all the key attributes in an event log.
    \item We propose a two-stage method to avoid the combinatorial explosion to reduce our computational complexity from $\mathcal{O}(N^3)$ to $\mathcal{O}(k^3)$ where $k \ll N$.
\end{enumerate}

The remainder of this paper is organized as follows. 
Section~\ref{sec:related_work} summarizes the prior art and the motivation of our work. 
Section~\ref{sec:proposed_method} describes the proposed method. 
Section~\ref{sec:performance_evaluation} shows the results of the performance evaluation. 
Section~\ref{sec:open_problems} discusses possible future work. Section~\ref{sec:conclusions} concludes the paper.

\section{Related Work}
\label{sec:related_work}
\subsection{Process Discovery}
\label{sec:process_discovery}
The mainstream research is the design of process discovery algorithms or miners. Process discovery is a method to extract the ordered sequences of processes from an event log. As it is infeasible to cover all the process discovery algorithms due to space limitations, we introduce some well-studied ones here. According to a recent survey paper~\cite{Garcia2019-az}, Cook and Wolf proposed the first process discovery algorithm based on a finite-state machine (FSM) in 1995~\cite{Cook1995-zg}. van der Aalst proposed to apply Petri-nets to process mining, which is often called Alpha miner, in 2002~\cite{Van_der_Aalst2002-xu}. Weijters and van der Aalst extended Alpha miner and proposed a new algorithm called heuristics miner (HM) to handle noisy event logs by taking into account the frequency of dependent activities~\cite{Weijters2006-yz}. It also handles possible short loops that cannot be dealt with in Alpha miner. HM is often chosen as it has been proven to be successful in real-event logs (e.g.,~\cite{De_Weerdt2012-so}). Hence, it is often used as a base miner as in~\cite{Vanden_Broucke2017-rp}. vanden Broucke and De Weerdt solved several issues in HM and proposed an extended algorithm called Fodina~\cite{Vanden_Broucke2017-rp}.
Leemans~et~al. proposed a series of inductive miners (IMs) \cite{Leemans2013-iv, Leemans2014-go, Leemans2015-jx}.
IM discovers a set of block-structured process models to ensure that the discovered models are sound and fit the observed behaviour~\cite{Leemans2013-iv}. Leemans~et~al. extended the original IM to filter out infrequent behavior quickly~\cite{Leemans2014-go}. They further extended IM to make it scalable by discovering a directly-follows graph (DFG) from an event log once and applying a divide-and-conquer strategy~\cite{Leemans2015-jx}. Augusto~et~al. proposed a split miner (SM) \cite{Augusto2021-us}. SM discovers a process model by filtering a DFG induced by an event log and identifying the combinations of split gateways that capture the concurrency, conflict, and causal relations between neighbors in the DFG.

\subsection{Pre-processing Raw Data}
The above process discovery methods can be applied to event logs with explicit case-ids, timestamps, and activities. 
However, raw data, which are often extracted from systems such as customer relationship management (CRM) and enterprise resource planning (ERP), are not yet ready for event logs. Diba~et~al. argue that we typically need to follow three steps to generate event logs from raw data, which are (i) event extraction, (ii) event correlation, and (iii) event abstraction~\cite{Diba2020review}. 

The event extraction is a process to retrieve raw event logs from databases. The event correlation is a step to correlate events by cases in the extracted logs. There could be cases where we need to correlate events (i) when raw data is missing case-ids, which correlate the series of events in a log, and (ii) when case-ids are hidden in raw data and unknown. The event correlation techniques have been well studied to solve the former issue. An efficient approach is key, as there could be many possible combinations for correlating events in such unlabelled raw data. For instance, Ferreira and Gillblad proposed a Markov-chain-based approach to infer a process model leveraging the activity transition patterns~\cite{Ferreira2009markov}.
Motahari-Nezhad~et~al. proposed efficient algorithms and heuristics to discover correlated events that could interest users~\cite{Motahari-Nezhad2011ec}. 
Bayomie~et~al. proposed a series of case-id inference algorithms with a complete process model (e.g.,~\cite{Bayomie2019-iy}\cite{Bayomie2022-nk}). Their approaches find the best event correlation via optimization. Specifically, they try to minimize the misalignment between a generated event log and an input process model and the activity execution time variance across cases using a simulated annealing algorithm. Helal and Awad proposed an online approach to handle streaming raw data~\cite{Helal2022online}. Reguieg~et~al. proposed a parallelized algorithm with MapReduce to scale the event correlation process~\cite{Reguieg2015scaling}.

For the cases when case-ids are hidden in raw data and unknown, few approaches have been investigated. 
Abbad Andaloussi~et~al. proposed an unsupervised case-id identification method~\cite{Abbad_Andaloussi2018-ur}. It identifies a case-id column in a given event log when it is not specified. Its idea is to find an attribute that seems most likely to be the case-id by applying a process discovery algorithm to every case-id candidate column and measuring its likeliness based on performance metrics such as recall and simplicity. Its rationality comes from the fact that a process model discovered with the correct case-id should give us better performance than that discovered with a wrong case-id column.
Bala~et~al. proposed a method to identify the case-id and activity columns in a raw event table~\cite{Bala2018-rb}. They leveraged the fact that we should see some repetitiveness of case-ids and activity columns. However, repetitiveness alone detects case-ids falsely. Hence, they considered the pairs of events with 
high individual repetitiveness but a low pairwise repetitiveness to reduce such a false detection.


\subsection{Motivation of Our Work}
Most of the existing process discovery algorithms require users to specify the key attributes in an event log, which would be problematic in some situations. First, it is tedious to manually look for these attributes as an event log is expected to contain tens of attributes. Furthermore, as process mining would be conducted not only by practitioners but also by laypeople, an automated approach is preferable, where a user does not have to specify attributes to be used as the key attributes.
Our goal is fully automated process mining, where no human intervention is required to discover a meaningful process flow from an event log.  

Abbad Andaloussi~et~al.'s work~\cite{Abbad_Andaloussi2018-ur} would be the closest to our goal. However, there are challenges to truly automated process mining. First, it only identifies a case-id attribute, not a timestamp or activity. Second, its computation is exhaustive as it repeats process discovery to every case-id candidate. If we are to identify case-id, timestamp, and activity attributes at once, the second challenge would be a big issue. Hence, our work aims to identify all the key attributes in an event log without incurring exhaustive computation.


\section{Proposed Method}
\label{sec:proposed_method}
\begin{figure*}[t]
    \centering
    \includegraphics[width=.95\linewidth]{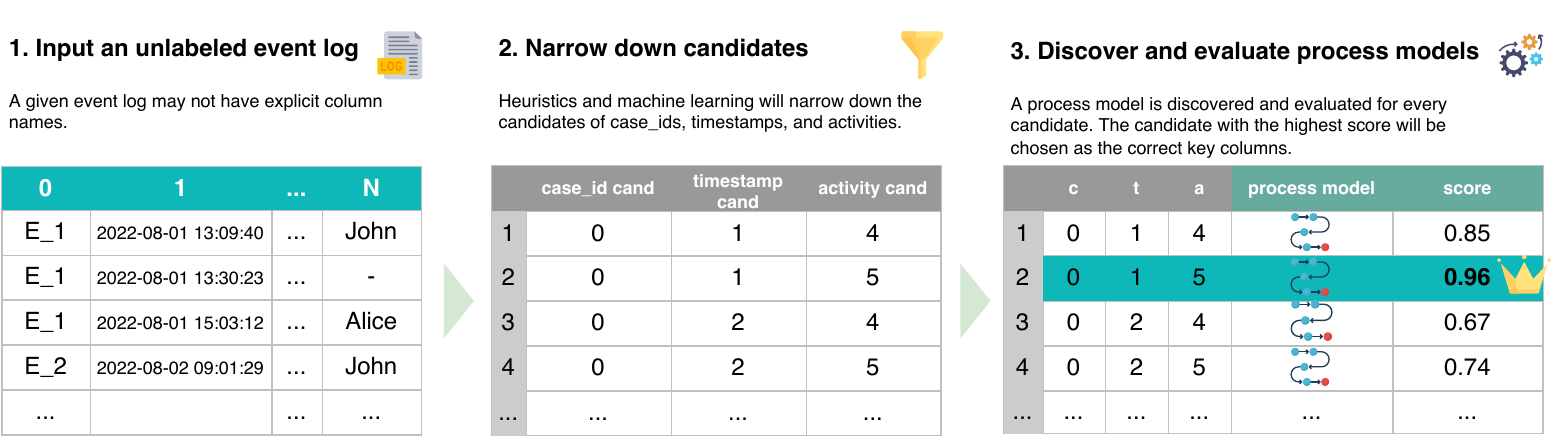}
    \caption{Overview of our method.}
    \label{fig:overview}
\end{figure*}
We propose a two-stage key attribute identification method. As an event log may have many attributes, we first narrow down candidates by analyzing the characteristics of values in each key attribute. If we could identify a single candidate for each key attribute, we output them as the key attributes. Otherwise, we identify the best candidate by evaluating every combination of key attribute candidates using a process discovery algorithm. We choose the candidate tuple with the highest score as the key attributes.

\figurename~\ref{fig:overview} illustrates our method. Our assumption on the input event log is that it surely contains the key attributes, but they are not explicitly given.

\subsection{Narrow down candidates}
\begin{figure*}
\begin{minipage}[c]{.27\linewidth}
    \centering
    \captionof{table}{Extracted features.}
    \label{tab:features}
    \resizebox{\linewidth}{!}{
    \begin{tabular}{ll}
        \toprule
        \textsc{Feature} & \textsc{Definition} \\
        \midrule
        $f_\text{s.letters}$ & Ratio of small letters \\
        $f_\text{l.letters}$ & Ratio of large letters \\
        $f_\text{digits}$ & Ratio of digits \\
        $f_\text{spaces}$ & Ratio of spaces (including tabs) \\
        $f_\text{symbols}$ & Ratio of symbols \\
        $f_\text{chars}$ & Number of characters \\
        $f_\text{words}$ & Number of words in a value \\
        \midrule
        $f_\text{r.unique}$ & Ratio of unique values \\
        $f_\text{m.unique}$ & Mean of number of each unique value \\
        \bottomrule
    \end{tabular}
    }
   \end{minipage}
   \hfill
   \begin{minipage}[c]{.67\linewidth}
    \centering
    \includegraphics[width=\linewidth]{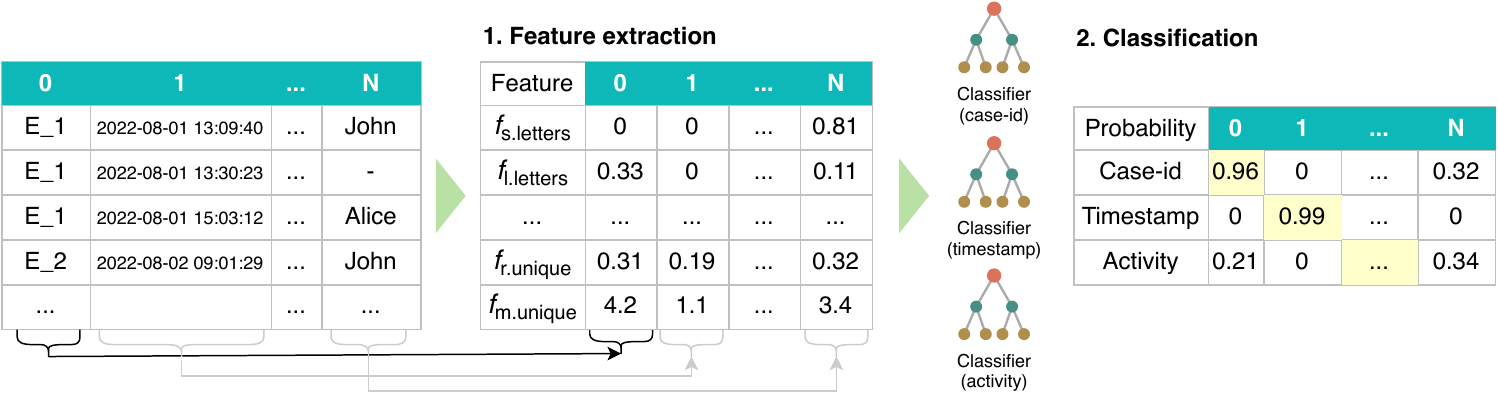}
    \caption{Overview of our first stage.}
    \label{fig:first_stage}
   \end{minipage}
\end{figure*}
We first narrow down the candidates of the key attributes against a given unlabeled event log. Figure~\ref{fig:first_stage} shows the overview of the procedures of this step. 

\subsubsection{Feature extraction}
We leverage the fact that there are noticeable characteristics in each key attribute. For instance, for case-ids, the same case-ids should appear multiple times in an attribute, otherwise, we cannot extract a process model with a single process. Similarly, for timestamps, we can expect many digits and symbols such as `-' and `:' in the values. To capture such characteristics, we extract the nine features listed in \tablename~\ref{tab:features} from each column.\footnote{Note that we only use the first $N_\text{rows}$, say 1,000, rows for feature extraction to reduce computation time.}
We call the first top seven features in the list local features and the last two global features. The local features are to extract characteristic features from each attribute value. For instance, for timestamps, we would expect more digits and symbols such as `-' and `:' than alphabets in the values. In contrast, the global features are to extract the frequency of the same values in a column. For instance, we expect the same values to appear multiple times for a case-id column.
As each local feature is a vector calculated from a value in an attribute, we calculate their mean values as the final local features. For instance, if the local features of an attribute contain three values, $\{$Case-01, Case-02, Case-03$\}$, as each value involves three small letters, i.e. `a', `s', and `e', out of six characters, $f_\text{s.letters} = (3/6 + 3/6 + 3/6) / 3 = 0.5$.

In contrast, global features are features extracted from values in an attribute. For instance, if values in an attribute are $\{1, 2, 3, 1, 2\}$, three unique values, namely 1, 2, and 3, appear twice, twice, and once, $f_\text{r.unique} = 3 / 5 = 0.6$ and $f_\text{m.unique} = (2 + 2 + 1) / 3 = 1.67$, respectively.

\subsubsection{Classification}
We then leverage supervised machine learning with the above features to narrow down the candidates for the key attributes. We chose supervised machine learning because many labeled event logs are available online.\footnote{E.g., \url{http://www.processmining.org/event-data.html}} We build three binary classifiers that output an attribute's probabilities of case-id, timestamp, and activity. For an unlabeled event log, we calculate the above features of each column and input them to trained classifiers to obtain the probabilities of whether an attribute belongs to case-id, timestamp, or activity. We choose the column(s) with the $N_\text{top}$ highest probability as candidates for each key attribute. The larger $N_\text{top}$ is, the more candidates we would obtain. For instance, when $N_\text{top} = 3$, we will at least have $3^3 = 27$ candidates. Hence, the possible $N_\text{top}$ would be one or two. We will evaluate the relationships between $N_\text{top}$, accuracy, and computation time in Section~\ref{sec:performance_evaluation}. If $N_\text{top} = 1$ and we have only one single column for each key attribute, we output them as the key attributes and do not proceed to the next stage. Otherwise, i.e., when more than one column has the highest probability, we need to identify the best one through the second stage, which we will explain in the next section.

\subsection{Discover and evaluate process models}
When having more than one candidate, we discover and evaluate a process model for each and determine the best one based on a performance metric. For an event log labeled by each candidate, we divide it into two parts by a case-id column (candidate) and evaluate the goodness of the chosen attributes with 2-fold cross-validation (CV). Specifically, we first use the first part for process model discovery and the second part for evaluation, and then swap the parts and do the same evaluation again. For instance, when a case-id candidate consists of $\{1, 2, \cdots, 9, 10\}$, we first discover the process model with the event log that contains $\{1, 2, \cdots, 5\}$ and evaluate it with the event log containing $\{6, 7, \cdots, 10\}$. We then swap and evaluate the training and evaluation event logs and calculate the average of these two scores as the final score. 

We can use any process discovery algorithms for this (e.g.,~\cite{Weijters2006-yz, Leemans2013-iv, Leemans2014-go}) and performance metrics (e.g., fitness, precision, generalization, simplicity, and their combinations~\cite{Buijs2014-zd, Abbad_Andaloussi2018-ur}). 
Fitness indicates how much of the observed behavior in an event log fits (or explains) the discovered process model. Precision, in contrast, quantifies how fewer unnecessary possible paths are generated by the discovered model. Generalization is to quantify how the discovered model has the flexibility to an unseen event log. Last but not least, structure or simplicity quantifies how fewer nodes are used in the discovered model.

In some cases, a process discovery algorithm takes a long time to discover and evaluate a process model or even fails to do so. This would not happen in the case where the combination of the key attributes is correct but would happen when it cannot find similarities among cases due to the wrong choice of the combination. To further save computation time, we introduce two thresholds, $th_\text{dis}$ and $th_\text{eval}$, to simply stop discovering or evaluating a process model and return 0 as a score after $th_\text{dis}$ or $th_\text{eval}$ has passed for process discovery or evaluation. These parameters have to be tuned based on the computer's specifications.

\section{Performance Evaluation}
\label{sec:performance_evaluation}
\begin{table*}[t]
    \begin{minipage}[t]{.2\linewidth}
    \centering
    \caption{Datsets used in our evaluation.}
    \label{tab:datasets}
    \resizebox{\linewidth}{!}{
    \begin{tabular}{lr}
        \toprule
        \textsc{Dataset} & \textsc{\#Columns} \\
        \midrule
        BPIC2011 & 128 \\
        BPIC2012 & 7 \\
        BPIC2013 incident management & 12 \\
        BPIC2013 problem management (closed) & 12 \\
        BPIC2013 problem management (open) & 11 \\
        BPIC2015 1 & 29 \\
        BPIC2015 2 & 28 \\
        BPIC2015 3 & 29 \\
        BPIC2015 4 & 29 \\
        BPIC2015 5 & 29 \\
        BPIC2017 & 19 \\
        BPIC2018 & 75 \\
        BPIC2019 & 21 \\
        BPIC2020 & 22 \\
        \bottomrule
    \end{tabular}}
    \end{minipage}
    \hfill
    \begin{minipage}[t]{.38\linewidth}
        \caption{Parameters of the process discovery algorithms.}
        \label{tab:parameters}
        \resizebox{\linewidth}{!}{
        \begin{tabular}{clr}
            \toprule
            \textsc{Miner} & \textsc{Parameter} & \textsc{Value}\\
            \midrule
            IM & Noise threshold & 0.2\\
            \midrule
            HM & Dependency threshold & 0.5\\
            HM & AND measure threshold & 0.65\\
            HM & Minimum number of occurrences of activity & 1\\
            HM & Minimum DFG occurrences & 1\\
            HM & DFG pre-cleaning noise threshold & 0.05\\
            HM & Loop length two threshold & 2\\
            \bottomrule
        \end{tabular}}
    \end{minipage}
    \hfill
    \begin{minipage}[t]{.3\linewidth}
    \centering
    \caption{Breakdown comparison of computation time.}
    \label{tab:detailed_computation_time}
    \resizebox{\linewidth}{!}{
    \begin{tabular}{ccrr}
        \toprule
        \multirow{2}{*}{$k$} & \multirow{2}{*}{\textsc{Miner}} & \multicolumn{2}{c}{Computation time [s]} \\
        \cmidrule{3-4}
         & & \textsc{Stage} 1 & \textsc{Stage} 2 \\
        \midrule
        1 & HM & 0.58 & 5.65 \\
        1 & IM & 0.58 & 17.15 \\
        2 & HM & 0.60 & 17.01 \\
        2 & IM & 0.60 & 43.68 \\
        \bottomrule
    \end{tabular}}
    \end{minipage}
\end{table*}
We conducted a performance evaluation to clarify the validity of our method.
We evaluated accuracy and computation time against the recent business process intelligence challenge (BPIC) datasets listed in \tablename~\ref{tab:datasets}. We used the BPIC datasets that explicitly contain the key attributes (i.e., ``case:concept:name'' for case-id, ``time:timestamp'' for timestamp, and ``concept:name'' for activity).\footnote{BPIC 2014 and 2016 datasets do not contain these attributes, and we thus did not use them.}\footnote{Although we used the explicitly labeled key attributes as unique correct tuples for the performance evaluation purpose, other tuples could benefit users. From this point of view, our first step may not be sufficient for the evaluation purpose where there is an assumption that only a single correct set of tuples exists. However, in reality, there could be more than one set of key attributes on the table that could be used as an activity, a case-id, or a timestamp~{\cite{Motahari-Nezhad2011ec}}, and we are technically not able to identify one tuple.} Accuracy was measured for each key attribute and defined as the number of correctly identified cases divided by the number of trials. Computation time was measured by the stages. As the first stage includes supervised machine learning, we executed a hold-out validation where one dataset is kept for testing while the others are used for training. We evaluated the accuracy and computation time of each dataset 10 times each. 
Our method is miner-agnostic, meaning any mining algorithm can be used, and thus we referred to Augusto~et~al.'s seminal benchmark paper~\cite{Augusto2019-cb} to choose the miners for evaluation. According to \cite{Augusto2019-cb}, although there is no dominant miner, inductive miner (IM) is found to be one of the most accurate miner. In contrast, heuristics miner (HM) is not the best or the latest algorithm, but we would like to see how the difference in miners affects the final result of our method. Hence, we tested the two process discovery algorithms, IM~{\cite{Leemans2013-iv}} and HM~{\cite{Weijters2006-yz}}, in the second stage, and their parameters are listed in \tablename~\ref{tab:parameters}, which are all default parameter settings defined by their libraries.
We implemented and evaluated our method using Python and its process mining library, PM4Py~\cite{Berti2019-xi}.\footnote{Our source code including dataset preparation, evaluation, and result summarization is available at \url{https://github.com/kentaroh-toyoda/research-process-mining}.}
We extracted the first $N_\text{rows} = 1,000$ rows for the evaluation for each dataset. We also set $th_\text{dis} = 5$ seconds and $th_\text{eval} = 60$ seconds.
The reason behind the choice of $N_\text{rows}$ is that we wanted to streamline the condition of the evaluation and to make sure the computation finishes in a reasonable time. Likewise, $th_\text{dis}$ and $th_\text{eval}$ were heuristically chosen so that we did not miss the correct combination of key attributes.

\begin{figure*}[t]
  \centering
  \begin{minipage}[t]{0.33\linewidth}  
    \centering
    \includegraphics[width=\linewidth]{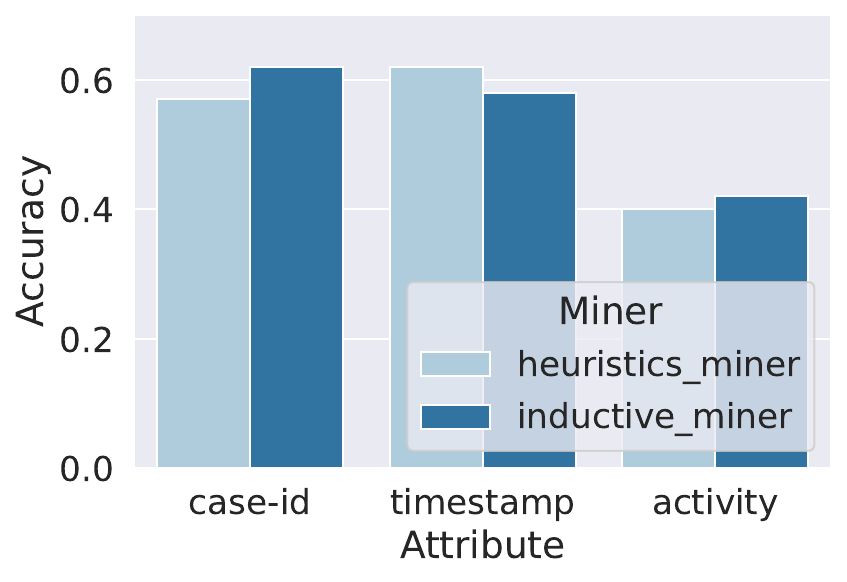}
    \caption{Accuracy versus miners.}
    \label{fig:accuracy_miners}
  \end{minipage}%
  \hfill  
  \begin{minipage}[t]{0.33\linewidth}  
    \centering
    \includegraphics[width=\linewidth]{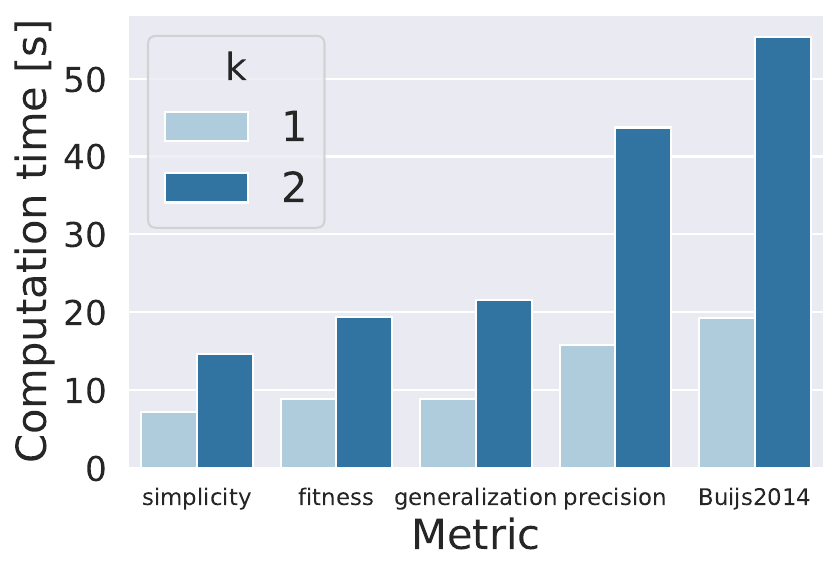}
    \caption{Computation time versus metrics.}
    \label{fig:computation_time_metrics}
  \end{minipage}%
  \hfill  
  \begin{minipage}[t]{0.33\linewidth}  
    \centering
    \includegraphics[width=\linewidth]{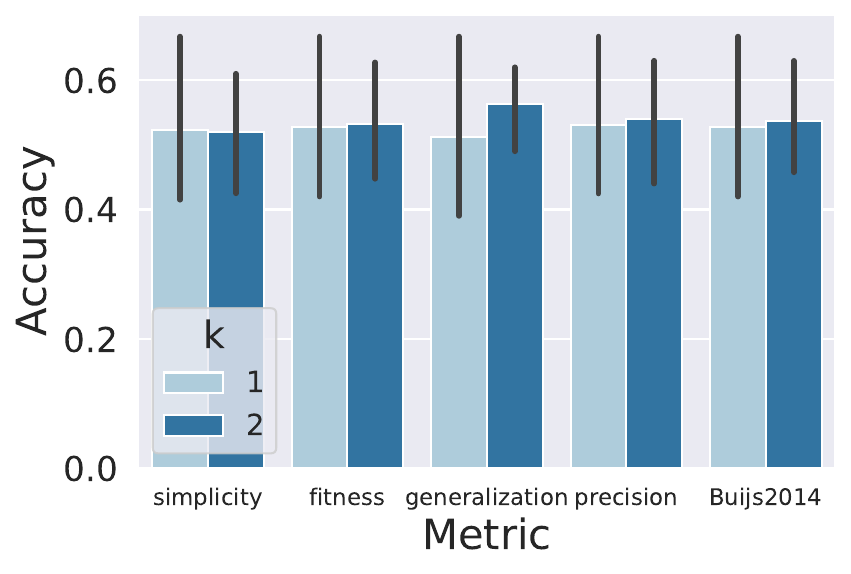}
    \caption{Accuracy versus metrics.}
    \label{fig:accuracy_metrics}
  \end{minipage}
\end{figure*}
Let us first discuss how miners used in the second stage affect the accuracy and computation time. \figurename~\ref{fig:accuracy_miners} and \tablename~\ref{tab:detailed_computation_time} show the accuracy and computation time to compare miners.
As can be seen from \figurename~\ref{fig:accuracy_miners}, we do not have a miner that dominates the other. When we average the three types of accuracy, IM is slightly better than HM; however, it is not significant. However, when we compare computation time by the miners, as in \tablename~\ref{tab:detailed_computation_time}, the difference is significant. HM is 2-3 times faster than IM regardless of $k$. As can be seen from the table, the computationally dominant part resides in the second stage to discover a process model and evaluate it. We are interested in how the metric used in the evaluation part of the second stage affects the computation time and accuracy. \figurename~\ref{fig:computation_time_metrics} shows the computation time by the metrics. Buijs~et~al.'s weighted average of the four individual metrics~\cite{Buijs2014-zd}, which we call Buijs2014, is the most time-consuming because it computes all four metrics. Among the four fundamental metrics, we found that precision is the most time-consuming metric. When we look at the accuracy results in \figurename~\ref{fig:accuracy_metrics}, we cannot see a significant difference in terms of accuracy, but generalization achieves the best accuracy among the five metrics. From these results, we can say that generalization is the most balanced metric in the second stage.
\begin{figure*}[t]
  \centering
  \begin{minipage}[t]{0.33\linewidth}  
    \centering
    \includegraphics[width=.95\columnwidth]{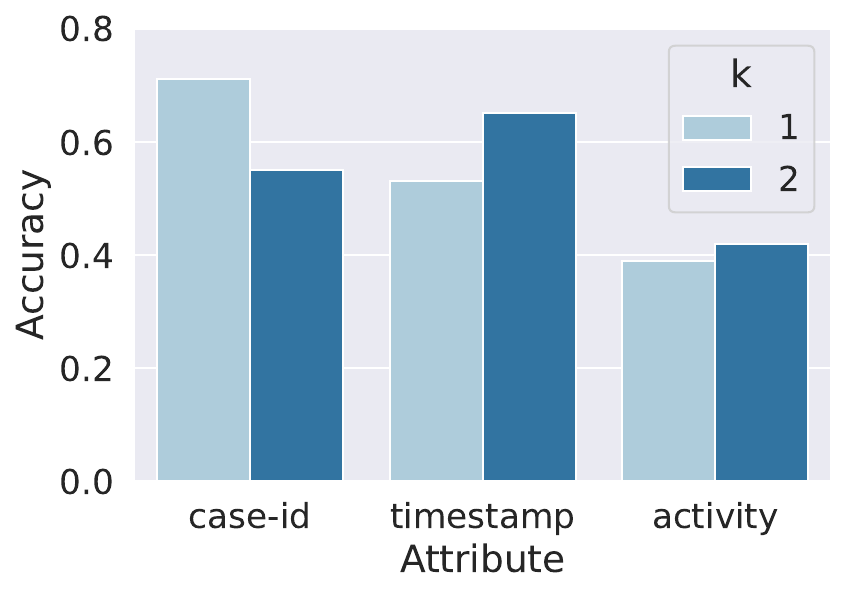}
    \caption{Accuracy versus attributes.}
    \label{fig:accuracy_attributes}
  \end{minipage}%
  \hfill  
  \begin{minipage}[t]{0.33\linewidth}  
    \centering
    \includegraphics[width=.95\columnwidth]{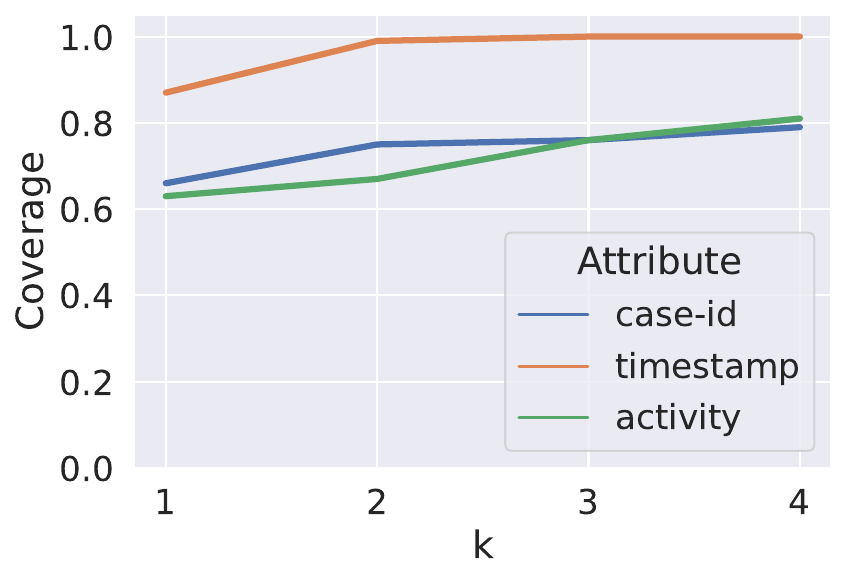}
    \caption{Coverage versus $k$.}
    \label{fig:coverage_k}
  \end{minipage}%
  \hfill  
  \begin{minipage}[t]{0.33\linewidth}  
    \centering
    \includegraphics[width=.95\columnwidth]{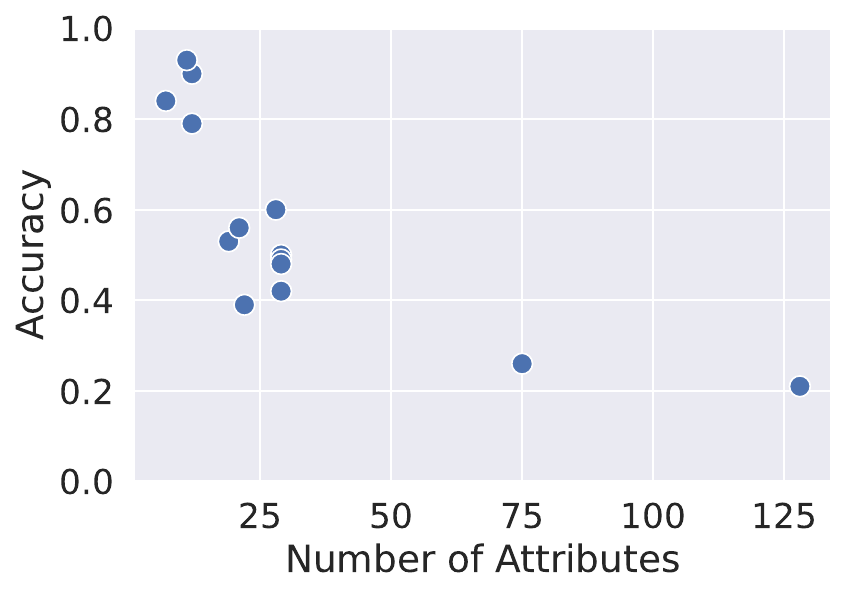}
    \caption{Accuracy versus the number of attributes in datasets.}
    \label{fig:accuracy_n_attributes}
  \end{minipage}
\end{figure*}
We then discuss how to choose $k$. Technically, the first stage should leave the correct key attributes. We evaluated a metric of coverage, whether the first step can successfully retain the correct key attributes. We measured the coverage metric by counting the cases where the correct key attributes were kept in the first step and dividing it by the total cases for $k = 1$ and $2$. We evaluated it against all the datasets and averaged the results. \figurename~{\ref{fig:coverage_k}} shows the results of coverage versus $k$. We can see from this figure that the larger $k$ the more correct key attributes we can retain. However, we filtered out the correct key attributes, in particular, case IDs and activities, even with $k = 4$.

The larger $k$, the more coverage we can achieve; however, from \tablename~\ref{tab:detailed_computation_time}, the larger $k$ the more computation time. This is because we need to evaluate more combinations of candidates for a large $k$. Hence, we want to set $k$ as small as possible. Luckily, we do not see much improvement when we increase $k$ from 1 to 2 as can be seen from \figurename~\ref{fig:accuracy_attributes}. 

Note that even if we set $k = 1$, meaning that when we try to leave only the single best candidate for each key attribute in the first stage, the second stage still does a job as can be seen from \tablename~\ref{tab:detailed_computation_time}. This means that we faced situations where there were multiple best candidates in the first stage.

Let us discuss the relationships between accuracy and the number of attributes in a dataset. \figurename~\ref{fig:accuracy_n_attributes} shows the averaged accuracy versus the number of attributes in the datasets. There is a negative correlation between accuracy and the number of attributes. We can explain this with a random guess where one randomly guesses which column should be used as case-id, activity, and timestamp. If we have 20 attributes, the accuracy of random guess would be roughly $0.05 (= 1 / 20)$ for each attribute. However, this decreases to 0.01 when a dataset contains 100 attributes. 

\section{Open Problems}
\label{sec:open_problems}
As the evaluation above shows, our method needs more polishment in terms of accuracy and computation time, in particular, when a given dataset is large. How well we can narrow down candidates in the first stage would be key to improving accuracy and reducing computation time at the same time. Our current approach is to judge the possibility of each key attribute against each table column. However, one possible solution is to take into account information captured from multiple columns or even from a whole table. Although their work's objective is not exactly the same as ours, Zhang~et~al. proposed to consider the relationships between adjacent columns to identify each table column's type~{\cite{zhang2020sato}}. Another promising direction would be to incorporate a context-aware approach (e.g., \cite{Mounira2010-rw, Song2019-fu, Vom_Brocke2021-gz}). If we could capture the context of the event log, then we would be able to find more suitable key attribute candidates. 

Regarding the computation time, one of the possible approaches is partially evaluating models with sampled event logs instead of using the whole one in the second stage. We will need to make sure how many samples are enough to assess the models. Another possibility to reduce computation time would be to polish our implementation. For instance, making use of a graphics processing unit (GPU) would help speed up the computation~\cite{Berti2022-qs}.

\section{Conclusions}
\label{sec:conclusions}
We have proposed a two-stage key attribute identification method for automated process discovery. If we were to evaluate all possible key attributes, their combinations would be too huge to try. To avoid such exhaustive computation, we first narrow down candidates by analyzing the characteristics of values in each column. If we could identify a single candidate for each key attribute, we output them as the key attributes. Otherwise, we identify the best candidate by evaluating every combination of key attribute candidates using a process discovery algorithm in the second stage. We chose the candidate tuple with the highest probabilities as the key attributes. We evaluated our method with 14 open datasets and showed that our method could feasibly identify key attributes for the datasets with fewer columns within a reasonable time. 

To the best of our knowledge, this paper is the first to identify the key attributes in an event log. Hence, we believe that the proposed method still has room for improvement. We have discussed the open problems and suggested several possible research directions toward fully reliable automated process discovery.

\section*{Acknowledgment}
This work was supported by the A*STAR ``Cyber-Physical Production System (CPPS) - Towards Contextual and Intelligent Response Research Program'' project with Grant No. A19C1a0018 and by the ``Distributed Smart Value Chain'' project with Grant No. M23L4a0001.

\ifCLASSOPTIONcaptionsoff
  \newpage
\fi

\bibliographystyle{IEEEtran}
\bibliography{ref}

\begin{IEEEbiography}[{\includegraphics[width=1in,height=1.25in,clip,keepaspectratio]{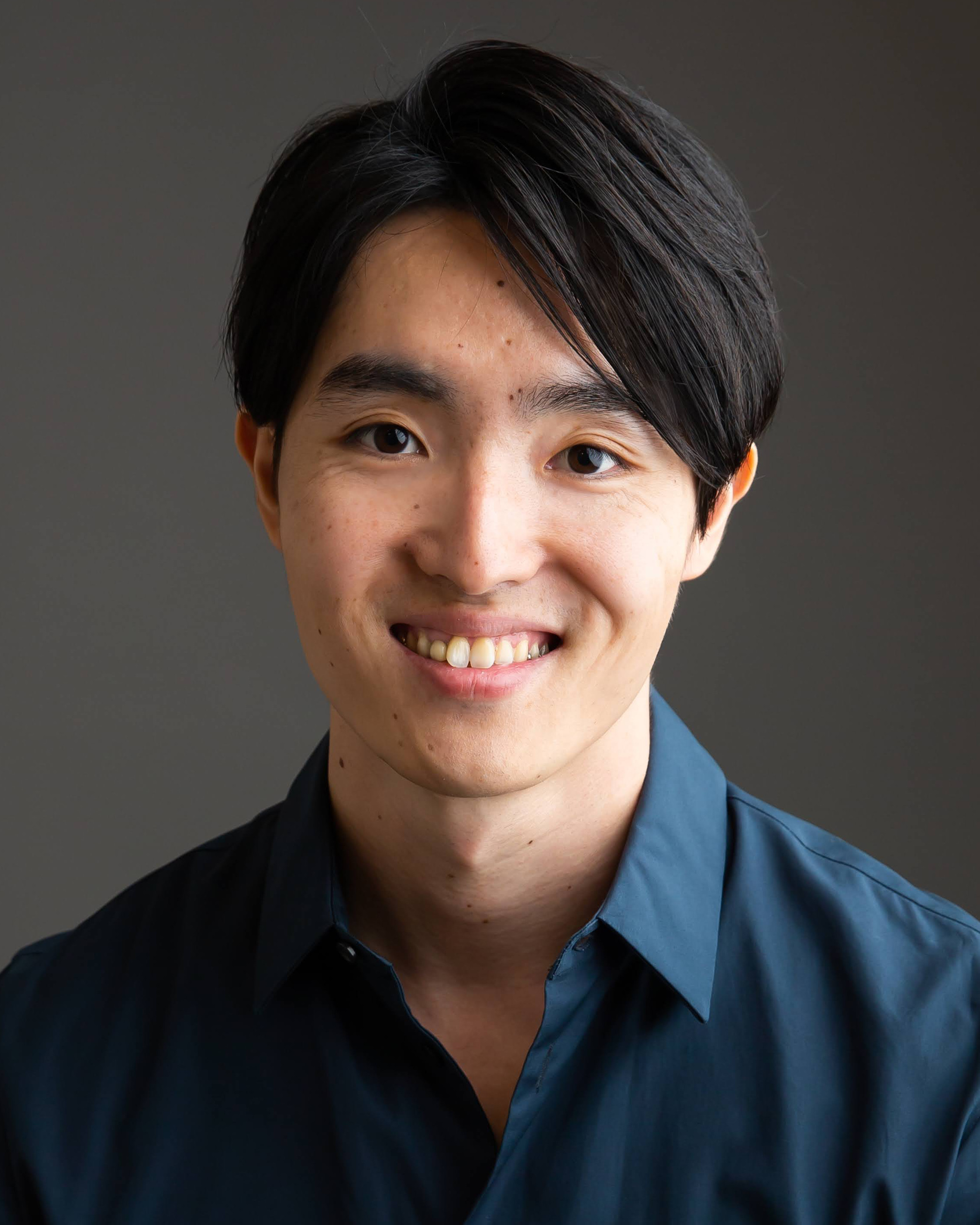}}]{Kentaroh Toyoda}
was born in Tokyo, Japan, in 1988. He received B.E., M.E., and Ph.D. (Engineering) degrees from the Department of Information and Computer Science, Keio University, Japan, in 2011, 2013, and 2016, respectively. He was a scientist at the Singapore Institute of Manufacturing Technology (SIMTech), A*STAR, from Apr. 2019 to Oct. 2022 and is currently a senior scientist at the Institute of High Performance Computing (IHPC), A*STAR, Singapore. His research interests include blockchain, security and privacy, data analysis, and mechanism design.
\end{IEEEbiography}

\begin{IEEEbiography}[{\includegraphics[width=1in,height=1.25in,clip,keepaspectratio]{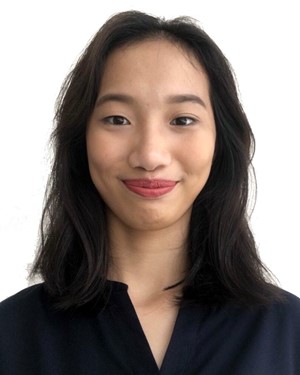}}]{Gan Kai Ying Rachel}
received a B.E. (Engineering) in Information Systems Technology and Design from the Singapore University of Technology and Design, Singapore in 2019. She was a research engineer at the Singapore Institute of Manufacturing Technology (SIMTech), A*STAR, from 2020 to 2022. She is pursuing her master's degree in AI at Nanyang Technological University (NTU), Singapore. Her research interests include data analytics, machine learning, and their applications.
\end{IEEEbiography}

\begin{IEEEbiography}[{\includegraphics[width=1in,height=1.25in,clip,keepaspectratio]{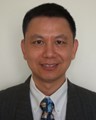}}]{Allan NengSheng Zhang}
is a Senior Pricipal Scientist with SIMTech, A*STAR, Singapore. He has more than 25 years of experience in the development of operations technologies (OT) using AI such as knowledge-based systems and enterprise information systems development. He is presently an Adjunct Associate Professor with the School of School of Mechanical and Aerospace Engineering, at Nanyang Technological University.  His research interests include knowledge management, data mining, machine learning, artificial intelligence, computer security, software engineering, software development methodology and standards, and enterprise information systems. He and his group are currently working toward research in cyber-physical system modeling, manufacturing system analyses including data mining, supply chain information management, supply chain risk and resilience management using a complex systems approach, multi-objective vehicle routing problems, and urban last-mile logistics.
\end{IEEEbiography}

\begin{IEEEbiography}[{\includegraphics[width=1in,height=1.25in,clip,keepaspectratio]{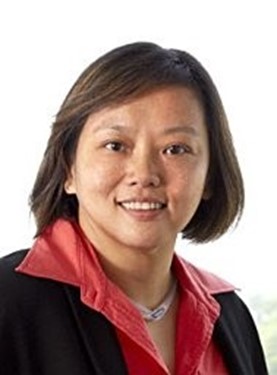}}]{Tan Puay Siew}
received a Ph.D. degree in computer science from the School of Computer Engineering, Nanyang Technological University, Singapore. She is presently an Adjunct Associate Professor with the School of Computer Science and Engineering, at Nanyang Technological University. In her full-time job at A*STAR, she leads the Digital Manufacturing Division of SIMTech and ARTC. Her research interests are in the cross-field disciplines of Computer Science and Operations Research for virtual enterprise collaboration, in particular, Distributed Smart Value Chains in the era of Industry 5.0.
\end{IEEEbiography}
\appendix
\section{}
{
\renewcommand{\arraystretch}{0.5}
\begin{table*}[th]
    \centering
    \caption{Accuracy by datasets, miners, and $k$. The best accuracy is highlighted in bold.}
    \label{tab:accuracy_datasets}
    \begin{tabular}{lccrrr}
    \toprule
    \multirow{2}{*}{\textsc{Dataset}} & \multirow{2}{*}{\textsc{Miner}} & \multirow{2}{*}{$k$} & \multicolumn{3}{c}{\textsc{Accuracy}} \\
    \cmidrule{4-6}
    & & & \textsc{Case-ID} & \textsc{Activity} & \textsc{Timestamp} \\
    \midrule
    BPIC2011 & NA & 1 & 0.00 & \textbf{0.29} & \textbf{1.00} \\
    \cmidrule{2-6}
     & \multirow{2}{*}{HM} & 1 & 0.00 & 0.00 & \textbf{1.00} \\
    & & 2 & 0.00 & 0.00 & 0.50 \\
    \cmidrule{2-6}
    & \multirow{2}{*}{IM} & 1 & 0.00 & 0.00 & \textbf{1.00} \\
    & & 2 & 0.00 & 0.00 & 0.45 \\
    \midrule
    BPIC2012 & NA & 1 & 0.70 & 0.90 & \textbf{1.00} \\
    \cmidrule{2-6}
     & HM & 2 & 0.48 & \textbf{1.00} & \textbf{1.00} \\
    \cmidrule{2-6}
    & IM & 2 & \textbf{0.80} & \textbf{1.00} & 0.71 \\
    \midrule
    BPIC2013 incident management & NA & 1 & \textbf{1.00} & 0.50 & \textbf{1.00} \\
    \cmidrule{2-6}
     & HM & 2 & 0.52 & 0.78 & 0.90 \\
    \cmidrule{2-6}
    & IM & 2 & 0.76 & \textbf{0.83} & 0.90 \\
    \midrule
    BPIC2013 problem management (closed) & NA & 1 & 0.50 & 0.80 & \textbf{1.00} \\
    \cmidrule{2-6}
     & HM & 2 & \textbf{0.98} & 0.82 & \textbf{1.00} \\
    \cmidrule{2-6}
    & IM & 2 & 0.76 & \textbf{0.89} & \textbf{1.00} \\
    \midrule
    BPIC2013 problem management (open) & NA & 1 & \textbf{1.00} & 0.90 & \textbf{1.00} \\
    \cmidrule{2-6}
    & HM & 2 & 0.88 & 0.90 & \textbf{1.00} \\
    \cmidrule{2-6}
    & IM & 2 & 0.82 & \textbf{0.95} & \textbf{1.00} \\
    \midrule
    BPIC2015 1 & NA & 1 & \textbf{1.00} & 0.00 & 0.00 \\
    \cmidrule{2-6}
    & \multirow{2}{*}{HM} & 1 & \textbf{1.00} & 0.48 & 0.00 \\
    & & 2 & 0.96 & \textbf{0.58} & 0.00 \\
    \cmidrule{2-6}
    & \multirow{2}{*}{IM} & 1 & \textbf{1.00} & 0.50 & 0.00 \\
    & & 2 & \textbf{1.00} & 0.50 & 0.00 \\
    \midrule
    BPIC2015 2 & NA & 1 & 0.67 & 0.00 & \textbf{1.00} \\
    \cmidrule{2-6}
    & \multirow{2}{*}{HM} & 1 & \textbf{0.86} & \textbf{0.43} & \textbf{0.86} \\
    & & 2 & 0.69 & 0.35 & 0.83 \\
    \cmidrule{2-6}
    & \multirow{2}{*}{IM} & 1 & \textbf{0.86} & \textbf{0.43} & \textbf{0.86} \\
    & & 2 & 0.54 & 0.30 & 0.48 \\
    \midrule
    BPIC2015 3 & NA & 1 & \textbf{1.00} & 0.00 & \textbf{1.00} \\
    \cmidrule{2-6}
    & \multirow{2}{*}{HM} & 1 & \textbf{1.00} & \textbf{0.28} & \textbf{1.00} \\
    & & 2 & 0.56 & 0.08 & 0.52 \\
    \cmidrule{2-6}
    & \multirow{2}{*}{IM} & 1 & \textbf{1.00} & 0.25 & \textbf{1.00} \\
    & & 2 & 0.55 & 0.07 & 0.47 \\
    \midrule
    BPIC2015 4 & NA & 1 & \textbf{1.00} & 0.00 & 0.00 \\
    \cmidrule{2-6}
     & \multirow{2}{*}{HM} & 1 & \textbf{1.00} & 0.43 & 0.00 \\
    & & 2 & 0.72 & \textbf{0.45} & \textbf{0.22} \\
    \cmidrule{2-6}
    & \multirow{2}{*}{IM} & 1 & \textbf{1.00} & 0.44 & 0.00 \\
    & & 2 & 0.88 & 0.40 & 0.15 \\
    \midrule
    BPIC2015 5 & \multirow{2}{*}{HM} & 1 & \textbf{0.90} & 0.46 & 0.15 \\
    & & 2 & 0.48 & 0.41 & 0.00 \\
    \cmidrule{2-6}
    & \multirow{2}{*}{IM} & 1 & \textbf{0.90} & \textbf{0.66} & \textbf{0.18} \\
    & & 2 & 0.55 & 0.40 & 0.00 \\
    \midrule
    BPIC2017 & NA & 1 & 0.40 & 0.00 & \textbf{1.00} \\
    \cmidrule{2-6}
    & HM & 2 & 0.34 & \textbf{0.11} & \textbf{1.00} \\
    \cmidrule{2-6}
    & IM & 2 & \textbf{0.78} & 0.02 & \textbf{1.00} \\
    \midrule
    BPIC2018 & NA & 1 & 0.00 & 0.00 & \textbf{1.00} \\
    \cmidrule{2-6}
    & \multirow{2}{*}{HM} & 1 & 0.00 & 0.00 & \textbf{1.00} \\
    & & 2 & \textbf{0.01} & 0.00 & 0.73 \\
    \cmidrule{2-6}
    & \multirow{2}{*}{IM} & 1 & 0.00 & 0.00 & \textbf{1.00} \\
    & & 2 & \textbf{0.01} & \textbf{0.01} & 0.69 \\
    \midrule
    BPIC2019 & NA & 1 & 0.25 & \textbf{0.62} & \textbf{1.00} \\
    \cmidrule{2-6}
    & \multirow{2}{*}{HM} & 1 & 0.00 & 0.00 & \textbf{1.00} \\
    & & 2 & 0.57 & 0.14 & \textbf{1.00} \\
    \cmidrule{2-6}
    & \multirow{2}{*}{IM} & 1 & 0.00 & 0.10 & \textbf{1.00} \\
    & & 2 & \textbf{0.69} & 0.18 & \textbf{1.00} \\
    \midrule
    BPIC2020 & NA & 1 & 0.00 & \textbf{0.33} & \textbf{1.00} \\
    \cmidrule{2-6}
    & \multirow{2}{*}{HM} & 1 & 0.00 & 0.25 & \textbf{1.00} \\
    & & 2 & 0.00 & 0.24 & 0.85 \\
    \cmidrule{2-6}
    & \multirow{2}{*}{IM} & 1 & 0.00 & 0.25 & \textbf{1.00} \\
    & & 2 & 0.00 & 0.32 & 0.85 \\
    \bottomrule
    \end{tabular}
\end{table*}
}
\tablename~\ref{tab:accuracy_datasets} shows each attribute's accuracy by the datasets for comparison. NA in the miner denotes the cases where we could identify the key attributes in the first stage.

\end{document}